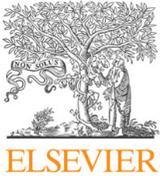

# Joint Max Margin and Semantic Features for Continuous Event Detection in Complex Scenes

Iman Abbasnejad[a,b], Sridha Sridharan[b], Simon Denman[b], Clinton Fookes[b], Simon Lucey[b]

[a]*Image and Video Laboratory, Queensland University of Technology (QUT), Brisbane, 4000, QLD, Australia*
[b]*The Robotics Institute, Carnegie Mellon University, 5000 Forbes Ave, PA, USA*

## ABSTRACT

In this paper the problem of complex event detection in the continuous domain (i.e. events with unknown starting and ending locations) is addressed. Existing event detection methods are limited to features that are extracted from the local spatial or spatio-temporal patches from the videos. However, this makes the model vulnerable to the events with similar concepts e.g. "Open drawer" and "Open cupboard". In this work, in order to address the aforementioned limitations we present a novel model based on the combination of semantic and temporal features extracted from video frames. We train a max-margin classifier on top of the extracted features in an adaptive framework that is able to detect the events with unknown starting and ending locations. Our model is based on the Bidirectional Region Neural Network and large margin Structural Output SVM. The generality of our model allows it to be simply applied to different labeled and unlabeled datasets. We finally test our algorithm on three challenging datasets, "UCF 101-Action Recognition", "MPII Cooking Activities" and "Hollywood", and we report state-of-the-art performance.

## 1. Introduction

Complex event detection is a challenging problem and has received increasing attention from computer vision researchers due to its potential in a number of applications such as human computer interaction, multimedia communication and video surveillance. Although many methods have been proposed in the literature the challenges have not been fully addressed yet. The majority of existing approaches are targeted toward the classification of pre-segmented events, rather than their detection and classification in a continuous video stream Zha et al. (2015); Yu and Yuan (2015). On the other hand, other methods that are mainly focused on continues problems Abbasnejad et al. (2015); Xu et al. (2014); Sun et al. (2015); Shou et al. (2016) fail to detect events with similar contexts such as "opening drawer" and/or "opening cupboard".

One popular strategy for event detection is to first extract visual features such as SIFT Lowe (2004) or SURF Bay et al. (2006) from the video frames, then pool or average over the entire video in order to represent it as a fixed dimensionality vector, and finally apply a linear classifier on it. Although these methods can perform remarkably well on atomic domains (i.e. videos with a simple event in a short duration video) they cannot perform efficiently on complex event problems (i.e. videos with several events and often last for a few minutes to even an hour) Xing and Yu (2015). One reason is that complex events consist of multiple intra-class variations and the pooling or averaging step, destroys all ordering and within-class variations among classes. In addition events are complex and are correlated and affected by different objects and actions. For instance, the event of "*cooking*" in a complex video may contain multiple objects or related actions such as "*knife*" and "*cutting*" or different locations such as "*kitchen*" or "*park*". In addition, detecting the events in a complex scene based on the local features extracted from each video frame is challenging and depending on the nature of the events different information may be required. For example to distinguish the events of "*walking*" and "*running*" temporal and body pose information are required, however to detect the events of "*Open drawer*" and "*Open cupboard*" information about the objects is needed.

One idea is to use a rich feature representation, i.e. Convo-lutional Neural Networks (CNN) of the input examples in order to improve the detection performance. Although the CNN architectures outperform other feature representation methods in many computer vision applications, in the area of event detection and temporal analysis, utilizing only CNN features for event detection could not be sufficient. However, it is not surprising, since human actions in video sequences are 3D spatiotemporal signals and analyzing events only based on the local features extracted from each video frame is insufficient Simonyan and Zisserman (2014a); Sun et al. (2015); Tang et al. (2012); Zha et al. (2015).

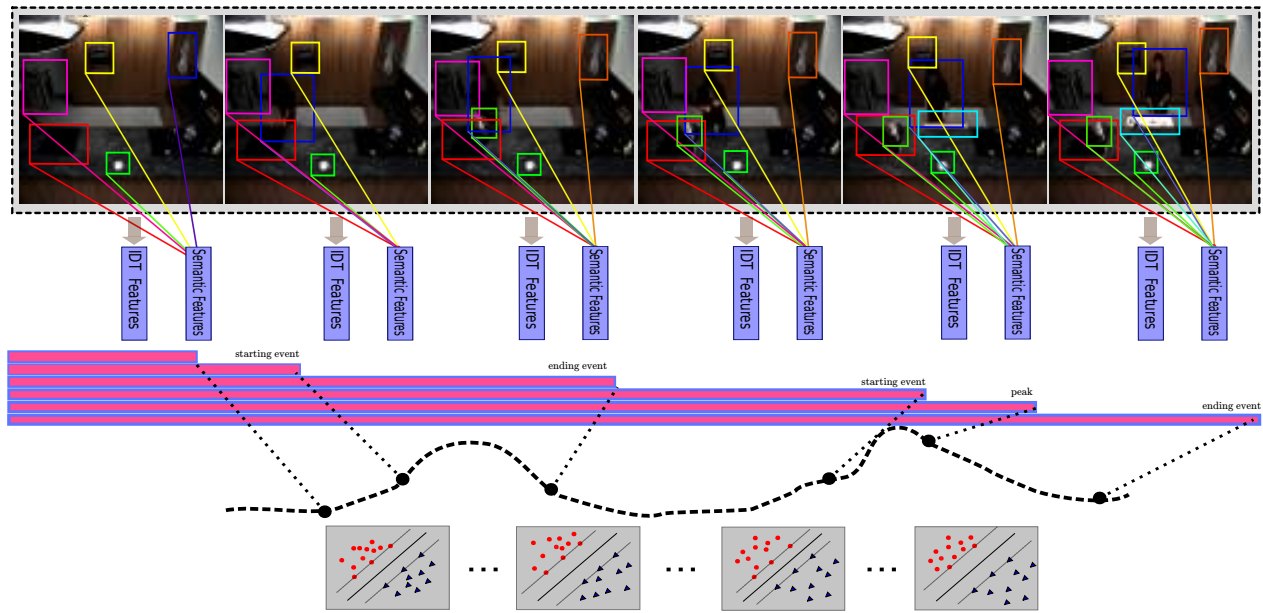

Fig. 1: Overview of our approach. For a sequence of video frames we first extract two sets of features: (i) Improved Dense Trajectories (IDT) and (ii) Semantic features from the video frames. The IDT features carry the temporal information about the video and the Semantic features carry the details about the objects and actions in each video frame (Section 3). We then feed the extracted features to a Structural Output SVM in order to learn the classifier (Section 4).

Recently it has been demonstrated that combining multiple features is an effective method for complex event detection and can improve the detection performance dramatically Natarajan et al. (2012b); Xu et al. (2014); Gan et al. (2016). For example, Natarajan et al. Natarajan et al. (2012b) demonstrated that combining multiple sets of features from different modalities such as audio and video-text improves the event recognition accuracy. Xu et al. Xu et al. (2014) showed that mixture of SIFT and Dense Trajectories features improves the detection performance. Gan et al. Gan et al. (2016) used web images as the semantic information to trim event videos and localize the relevant frames of videos in order to improve detection accuracy. Despite existing methods Xu et al. (2014); Gan et al. (2016) report state-of-the-art performance in event detection, one limitation with these approaches is that it is difficult to generalize them to any arbitrary video signal (i.e. silent videos or videos without text subtitles) therefore these methods are limited to the videos with only a few applications. Furthermore, most systems combine these features in a simple way without considering any correlations between different features for event detection, however they are correlated and complementary to each other.

In order to address the aforementioned limitations, we present a novel method to tackle the problem of efficient event detection. Our approach leverages the benefits of multiple sets of features in an adaptive framework for continuous event problems, i.e. detecting a particular event in an unknown sequence with unknown starting and ending locations. In this work, unlike the previous methods we utilize the features that are extracted from different aspects of the input videos (i.e. the semantic and temporal content) in order to capture the information about the sets of objects, actions and temporal details that are seen in each video frame. There are two benefits in this presentation, firstly utilizing semantic features provides information about the objects and actions that are seen in video frames and consequently improves the detection performance, secondly this presentation can be used to annotate videos with unknown labels.

To draw the correlations among the extracted features we present a new formulation on an adaptive framework based on a large margin structure. In particular, we generate an alternate frame-wise feature representation using the objects and actions that are seen in each frame, and learn an event detector on them that correctly classifies partially observed sequences. We also utilize our method in annotating and auto labeling unlabeled videos. By unlabeled videos we mean that we do not have any prior knowledge about the locations and labels of the events in the observed sequences. We shall discuss our proposed model in the subsequent sections.

**Paper Contributions:** In this paper the problem of continuous event detection in complex scenes is addressed. Recently it has been demonstrated that combination of different sets of features from different modalities (i.e. audio and motion) can improve detection performance Natarajan et al. (2012a). A drawback to these strategies, however, is that it is hard to generalize them to any arbitrary video sequence. Furthermore, previous methods combine multiple features in a simple way without considering any correlation among features. In this work we present a novel approach that leverages the benefits of multiple sets of features in an adaptive framework through a max-margin linear classifier. We transform the video data into a frame-wise representation that captures the objects and actions present in each video frame, and train a max margin classifier on the extracted features. There are two advantages for this formulation, (i) during training the classifier learns the correlations and dependencies among different objects, actions and events, and (ii) this presentation makes the learning method robust toward the small number of positive training examples since we put the weights on the margin with respect to the positive and negative training examples. We also use our method to train a classifier on sets of unlabeled training data for continuous event detection. Finally we evaluate our model on three challenging event datasets and demonstrate how our approach performs in comparison to the existing methods. Figure 1 visualizes an overview of our proposed model.

## 2. Event detection

Event detection refers to recognizing an event or events in a video. Different strategies have been developed in the literature in order to improve the detection performance Yang and Shah (2012); Tang et al. (2012); Liu et al. (2012); Natarajan et al. (2012a); Abbasnejad et al. (2015). One popular method is to learn a discriminative event detection function which is linearly applied to the observed data,

$$f(\mathbf{X}; \boldsymbol{\omega}) = \eta\{\mathbf{X}\}^T \boldsymbol{\omega} \qquad (1)$$

where $f(\mathbf{X}; \boldsymbol{\omega}) : \mathbb{R}^{M \times F} \to \mathbb{R}^1$ is a mapping function from the data domain to the output domain, the parameter $\boldsymbol{\omega} \in \mathbb{R}^{M \times 1}$ is the model parameter and $\eta\{\mathbf{X}\} \in \mathbb{R}^{M \times 1}$ is a vectorized feature representation of the multi-dimensional event sequence $\mathbf{X} \in \mathbb{R}^{M \times F}$; where, $M$ is the dimensionality of the signal and $F$ is the number of frames.

Although there are many advantages for maintaining a linear relationship between the data domain $\eta\{\mathbf{X}\}$ and the classifier Abbasnejad et al. (2015), there are still some drawbacks with this model: (i) the performance in this model is strongly influenced by the quality of the input features, (ii) decreasing the amount of training data reduces the classification accuracy, (iii) it fails to capture temporal information among the frames in the observed videos, (iv) due to intra-class variations it is hard to generalize such a method to the complex event problems, (v) since the filter, $\boldsymbol{\omega}$ has a fixed size in such a presentation, it cannot be applied for events with unknown starting and ending location.

Different heuristics have been proposed in the literature in order to tackle the aforementioned limitations. Chen et al. Chen et al. (2013) built a classifier based on the combination of Fisher vector coding representation and sliding windows techniques for event classification. Hoai et al. Hoai and De la Torre (2014) proposed to utilize the Structured Output SVM in conjunction with the Bag-of-words (BoW) framework to detect the sequential arrival events as early as possible. Izadinia et al. Izadinia and Shah (2012) assumed each complex video, can be interpolated as a mixture of some low-level features that can be treated as the hidden structures in a latent SVM model for modeling the complex events. Abbasnejad et al. Abbasnejad et al. (2015) used the idea of sliding windows via large margin classifiers to present a fast and computationally inexpensive method for continuous event detection. Yang et al Yang and Shah (2012) designed an approach based on the combination of multiple features from three different modalities (audio, scene and motion) to improve the detection performance.

On the other hand, with the progress of CNN networks in the other areas of computer vision, recent work in the field of event detection and temporal analysis has concentrated on applying these features to this task. However, as demonstrated by Simonyan and Zisserman (2014a); Sun et al. (2015) CNN features extracted from each video frame are insufficient for event detection. Several recent approaches in using CNN networks for action recognition have investigated the question of how to go beyond simply using the framewise appearance information and exploit the temporal information. A natural extension is to extend 2D ConvNets into time Ji et al. (2013) so that the first layer learns spatiotemporal features. In Karpathy et al. (2014), compared several approaches for temporal sampling, including early fusion (letting the first layer filters operate over frames similar to Ji et al. (2013)), slow fusion (consecutively increasing the temporal receptive field as the layers increase) and late fusion (merging fully connected layers of two separate networks that operate on temporally distant frames) Consequently through experiment they concluded that the network temporal modeling structure is not gaining much from the temporal information Feichtenhofer et al. (2016).

Alternatively the modeling of temporal dynamics can be achieved with more sophisticated deep learning models such as RNN Elman (1990) or LSTM Hochreiter and Schmidhuber (1997). These models have recently started to appear in the action recognition literature. In Baccouche et al. (2011), a 3D convolutional neural network followed by an LSTM classifier was successful at classifying simple actions. LSTMs have shown improved performance over a two-stream network for action recognition Donahue et al. (2015); Yue-Hei Ng et al. (2015). Recently, bi-directional LSTMs were also successful in skeletal action recognition Du et al. (2015). Sun et al. Sun et al. (2015) used LSTMs for action detection although their focus was on leveraging web images to help with video action detection. However, even after using LSTMs, deep learning methods perform only slightly better than fisher vectors built on handcrafted features for many action recognition tasks Yue-Hei Ng et al. (2015).

Although enormous progress has been made in action recognition and classification Rohrbach et al. (2015); Simonyan and Zisserman (2014a); Wang et al. (2015); Abbasnejad et al. (2015), the problem of event detection in the continuous domain (e.g. spatio-temporal localization of actions in longer videos) is still in progress. Recently the major focus in action detection has been on using and extracting low-level features from the videos. However, the extracted features are incapable of capturing the inherent semantic information in an event. Furthermore, previous methods do not draw any dependencies between the actions and the corresponding objects. In addition, these methods cannot be applied to many complex event video datasets due to the small number of positive training examples in the video sequences. To solve these problems Souza et al. d. Souza et al. (2014) developed a graph method for representing and combining interactions of actors with objects in event sequences. In the presented graph, nodes represent the items of interest (objects, actors, actions, etc) and edges connect their interactions. Gan et al. Gan et al. (2016) utilized the web images in order to clean and denoise the event videos. They extracted the CNN features from the trimmed videos to train the LSTM network. However, the main limitation of these methods is that they cannot perform well with large videos containing multiple actions and complex interactions between actions Souza et al. (2015).

Drawing upon current success, we present our strategy for efficient event detection. We focus on combining multiple sets of features (that has been demonstrated in the literature to be efficient for event detection) which are extracted from temporal and static parts of videos. We claim that this presentation is robust for efficient event detection and in contrast to the previous methods, it can be applied to any arbitrary video dataset.

## 3. Feature representation

As mentioned in Section 2 a good feature representation is crucial for efficient event detection. Traditional approaches typically use information extracted from local spatial or spatio-temporal patches from the video to recognize events. However, although local information is vital for video analysis, depending on the nature of the actions using only local information is insufficient. For example in order to classify two events of "*walking*" and "*running*" the temporal information is important, however for the events of "*take-out drawer*" and "*take-out cupboard*" the information about the objects i.e. "*drawer*" and "*cupboard*" are key features. In this section, we introduce our feature representation method. For our presentation we extract two sets of features, *"Semantic features"* and *"Temporal features"* for modeling the video events. Semantic features, extract details about the objects and actions that are seen in each video frame and temporal features extract the temporal information from time series measurements in a low-dimensional feature space.

### 3.1. Semantic features

This section provides details about the semantic feature extraction method we use in this paper. The semantic features carry the information about the objects and actions present in each video frame and are extracted by transforming DCNN features (see Section 3.1.1) into a word2vec representation (see Section 3.1.2)

### 3.1.1. DCNN Feature Extraction

Given the input image $I$, we start by extracting frame-based features. We start with the CNN architecture using the Caffe toolkit and the model shared by Simonyan et al. Simonyan and Zisserman (2014b). The key insight in this model is that by using smaller convolution filters ($3 \times 3$) and very deep layers, significant improvement can be achieved on the ImageNet Detection Challenge. This configuration has 16 weight layers with the first 13 weight layers convolutional layers, five of which are followed by a max-pooling layer and the last three weight layers are fully connected. Given the input image $I$ we simply extract the vector $\eta_c\{I\} \in \mathbb{R}^{4096 \times 1}$ from its probability layer (the last layer).

### 3.1.2. Representing Frames Semantically:

This section provides details about the semantic features we extract from the frames. The extracted features carry the information about the set of objects and actions that are seen in each video frame. For feature extraction, we build our model on the work presented by Karapathy et al. Karpathy and Fei-Fei (2014) where the goal of this model is to describe the input image in a sentence. We choose this model because it is accurate and it performs well on variety of datasets and images.

An overview of this model is as follows: in order to generate the caption for the input image $I$, we first extract the top 19 regions using a Region Convolutional Neural Network (RCNN) Girshick et al. (2014) and represent the input image $I$ as a $4096 \times 20$ dimensional matrix, $\mathbf{P} \in \mathbb{R}^{4096 \times 20}$:

$$
\begin{aligned}
\mathbf{P} &= [\mathbf{p}_1, \mathbf{p}_2, \ldots, \mathbf{p}_{20}] \\
\mathbf{p}_i &= W_m \eta_c\{I_i\} \quad \forall \ i = 1, \ldots, 20
\end{aligned}
\tag{2}
$$

where $\mathbf{p}_{i-19}$ is the feature representation of the $i$-th region and $\mathbf{p}_{20}$ is the feature representation of the whole image $I$ using the representation in Section 3.1.1. The parameter $W_m$ is the RCNN parameter and is learned during training of RCNN.

After extracting the top 19 regions from the input image $I$, we apply a Bidirectional Region Neural Network (BRNN) Schuster and Paliwal (1997) on the input image representation $\mathbf{P}$ in order to generate the sentence:

$$
\eta_s\{I\} = \min_{\mathbb{I}} \|\mathbf{P} - \mathbf{Cb}\|_2^2
\tag{3}
$$
$$
\text{s.t.} \quad \|\mathbf{b}\|_0 \leq k
$$

where $\mathbf{b} \in \mathbb{R}^{K \times 1}$ is a $K$ dimensional vector containing all zeros except for $k$ entries Abbasnejad et al. (2017) in the word vocabularies $\mathbf{C} \in \mathbb{R}^{E \times K}$. After generating the sentence, since each sentence has different dimensionality, we simply apply a 4096-dimensional word2vec model in order to represent the sentence as a fixed dimensionality vector.

### 3.2. Temporal features

As presented in the literature, temporal information is significant for video and event analysis. Given the input signal $\mathbf{X} \in \mathbb{R}^{M \times F}$ we represent the temporal features as:

$$
\eta_T\{\mathbf{x}_t\} = \min_{\mathbf{e}} \|\phi_T\{\mathbf{x}_t\} - \mathbf{Qe}_t\|_2^2 + \lambda \|\mathbf{e}_t\|_1
\tag{4}
$$

where in this work $\phi_T\{.\} \in \mathbb{R}^{N \times 1}$ belongs to the set of Improved Dense Trajectory (IDT) Wang and Schmid (2013) transformations which are applied to each video frame $\mathbf{x}_t$, $\mathbf{Q} \in \mathbb{R}^{N \times K}$ is the codebook and $\lambda$ is a parameter controlling the sparsity penalty. This algorithm is robust against camera motion and efficient for temporal feature extraction. The set of dense trajectories are obtained by tracking the points with median filter and different descriptors Wang et al. (2013): trajectory shape, HOG (Histogram of Oriented Gradients), HOF (Histograms of Optical Flow) and MBH (Motion Boundary Histogram), in five different scales. More details can be found on Wang and Schmid (2013).

### 3.2.1. Segment representation

As suggested in the literature, the traditional way to represent the video segments is simply averaging among all the frames and representing the whole video segment as a fix dimensionality vector. This representation eliminates the effects of the large number of negative training examples as presented in Ding et al. (2013). In this work we use this representation and apply it to the video frames as follows:

$$\psi_T(\mathbf{X}_{[1:f]}) = \frac{1}{f} \sum_{i=1}^{f} \eta_T\{\mathbf{x}_f\} \tag{5}$$

where $\psi_T(\mathbf{X}_{[1:f]}) \in \mathbb{R}^{K \times 1}$ is the temporal feature representation of the subsequence of time series $\mathbf{X}$ from the first frame to the $f$-th frame.

## 4. Event detection using multiple features

In this section, we provide our formulation for efficient event detection based on a large margin classifier. In particular, we create a framewise feature representation of input data by using the objects, actions and temporal features that are captured in each observed frame and train a classifier on top of them in a max-margin framework.

### 4.1. Learning SO-SVM with multiple features

Let $\mathbf{X}^i = [\mathbf{x}_1^i, \ldots, \mathbf{x}_f^i, \ldots, \mathbf{x}_F^i]$ be the $i$-th time series training example with its corresponding output labels $\mathbf{y}^i = [y_1^i, \ldots, y_f^i, \ldots, y_F^i]$. The goal of SO-SVM is to learn a mapping function $\omega$ from the space of training example, $\mathbf{X} \in \mathbb{R}^{M \times F}$ to the label classes $\mathbf{y} \in [-1, +1]$. The cost function for training SO-SVM can be written as Tsochantaridis et al. (2005):

$$\min_{\omega, b, \zeta^i} \frac{1}{2}\|\omega\|^2 + \frac{C}{n} \sum_{i=1}^{n} \zeta^i, \tag{6}$$

$$\text{s.t.} \quad \omega^T \psi(\mathbf{x}_f^i, \mathbf{y}_f^i) \geq \omega^T \psi(\mathbf{x}_f^i, \mathbf{y}^i) - \zeta^i$$

$$\forall i, \forall f = 1, \ldots, l^i$$

where $\zeta^i$ is the slack variable, $\mathbf{x}_f^i$ is the $f$-th training example and $n$ is the number of training examples. This formulation is convex and efficient for problems with structured output properties, such as sequences or graphs. Therefore, we present our formulation based on SO-SVM as follows:

$$\min_{\omega, b, \zeta^i} \frac{1}{2}\|\omega\|^2 + \frac{C}{n} \sum_{i=1}^{n} \zeta^i, \tag{7}$$

$$\text{s.t.} \quad \omega^T \psi_T(\mathbf{X}_f^i) \geq$$

$$\omega^T \psi_T(\mathbf{X}_{\mathbf{y}^i}^i) + \mu(\mathbf{y}_f^i, \hat{\mathbf{y}}_f^i) - \frac{\zeta^i}{\Delta(\mathbf{y}_f^i, \mathbf{y}^i)} \tag{8}$$

$$\forall i, \forall f = 1, \ldots, l^i$$

in this formulation, Eq.8 denotes that for each arrival frame in time $f$, the score of current frame $\omega^T \psi_T(\mathbf{X}_f^i)$ is required to be greater than the score of the events which have been seen from the first to $f$-th frames, $\omega^T \psi_T(\mathbf{X}_{\mathbf{y}^i}^i)$, by $\mu(\mathbf{y}_f^i, \hat{\mathbf{y}}_f^i)$. We define the score function as:

$$\mu(\mathbf{y}_f^i, \hat{\mathbf{y}}_f^i) = |\mathbf{y}_f^i - \hat{\mathbf{y}}_f^i|, \quad \hat{\mathbf{y}}_f^i = \omega^T \psi_S(\mathbf{X}_{\mathbf{y}_f^i}^i) \tag{9}$$

where $\mathbf{y}_f^i$ and $\hat{\mathbf{y}}_f^i$ are scores of the outputs with respect to the temporal and semantic features extracted from each video frame in time $f$ and $\psi_S(\mathbf{X}_{\mathbf{y}_f^i}^i)$ is:

$$\psi_S(\mathbf{X}_{\mathbf{y}_f^i}^i) = [\eta_S\{\mathbf{x}_1\}, \eta_S\{\mathbf{x}_2\}, \ldots \eta_S\{\mathbf{x}_f\}]$$

This presentation makes the classifier to put the weights on the margin with respect to the semantic details extracted from each frame. In other words, we change the margin with respect to the semantic features extracted from each video frame. We also put a rescaling parameter on the slack variable $\zeta^i$ and define it as:

$$\Delta(\mathbf{y}_f^i, \mathbf{y}^i) = |\mathbf{y}_f^i - \mathbf{y}^i| \tag{10}$$

This weight rescales the parameter $\zeta^i$ with respect to the correct detection at time $f$. This forces SO-SVM to give more more emphasis to the newly arrived frames and penalizes the margin constraint by a large amount for the events that are far from the event of interest (i.e. at the far end of the temporal window).

In order to see what we are optimizing in Eq. 7 we follow the work presented by Tsochantaridis et al. Tsochantaridis et al. (2005) who showed for the pairs of examples $\{\mathbf{X}, \mathbf{y}\}$ generated from some distribution $P(\mathbf{X}, \mathbf{y})$, the loss of the detector $g(.)$ is:

$$\mathcal{R}_{true}^{\Delta}(g) = \int_{\mathbf{X} \times \mathbf{y}} \Delta(\mathbf{y}, g(\mathbf{X})) dP(\mathbf{X}, \mathbf{y})$$

and since $P$ is an unknown distribution over the input examples therefore the performance of $g(.)$ is described by the empirical risk from the training data $\{\mathbf{X}, \mathbf{y}\}$. The upper bound for the empirical risk $\mathcal{R}_{emp}^{\Delta}$ can be defined from proposition 1.

**Proposition 1.** *The optimal solution for the SO-SVM is denoted by $(\omega^*, \zeta^*)$. Then the upper bound on the empirical risk for a set of training examples is:*

$$\mathcal{R}_{emp}^{\Delta}(\omega^*) = \frac{1}{n} \sum_{i=1}^{n} \zeta^{*i}$$

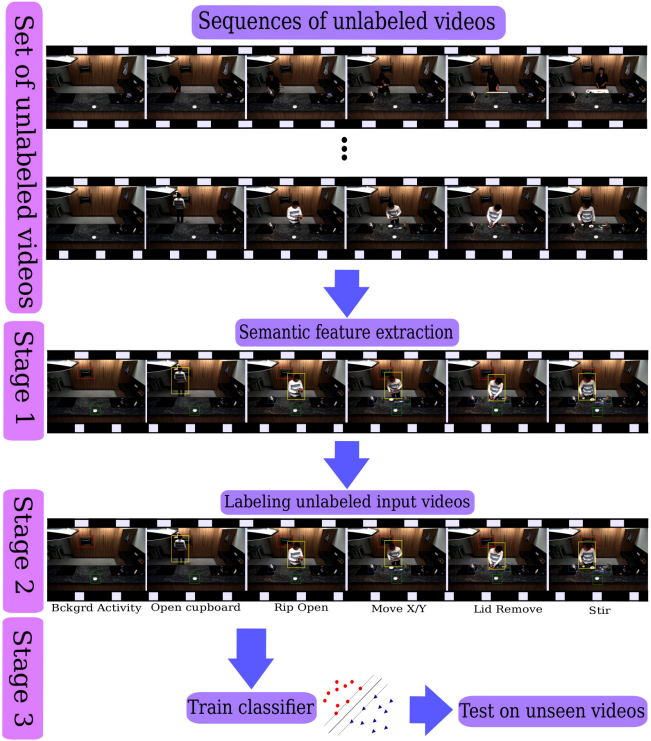

Fig. 2: Event detection on unlabeled video. Given a set of unlabeled videos we first apply our semantic feature extraction technique on each video frame (Stage 1). This gives us the set of frame labels (Stage 2). Finally we train and test our classifier on the unseen videos (Stage 3).

where for the formulation presented in Eq. 7 the optimal solution for the slack variable is:

$$\zeta^{*i} = \max_f \{0, \Delta(\mathbf{y}_f^i, \mathbf{y}^i)(\mu(\mathbf{y}_f^i, \hat{\mathbf{y}}_f^i) - \mathcal{F}(\boldsymbol{\omega}, \psi_T(\mathbf{X}^i))\}$$

where $\mathcal{F}(\boldsymbol{\omega}, \psi_T(\mathbf{X}^i)) = \boldsymbol{\omega}^T(\psi_T(\mathbf{X}_{\mathbf{y}^i}^i) - \psi_T(\mathbf{X}_{\mathbf{y}_f^i}^i))$. Therefore the upper bound on the empirical risk function for our formulation can be defined as:

$$\frac{1}{n} \sum_{i=1}^{n} \max_f \{ \Delta(\mathbf{y}_f^i, \mathbf{y}^i)(\mu(\mathbf{y}_f^i, \hat{\mathbf{y}}_f^i) - \mathcal{F}(\boldsymbol{\omega}, \psi_T(\mathbf{X}^i)) \} \quad (11)$$

Eq. ?? shows how the $f$-th arrival frame is being modeled using the temporal and semantic features extracted from the $f$-th frame. More precisely Eq. ?? demonstrates that our classifier learns the related correlation among the extracted objects and temporal features in the observed sequence.

### 4.2. Event detection on unlabeled video

As explained earlier one of the contributions of this paper is to present a method that is able to recognize events in an unlabeled video. By unlabeled video we mean that we do not have any information about the events, their locations and their labels in the observed video in the both training and testing phases. In this paper unlike the previous methods that utilize human resources to annotate the video frames, we present a method which is able to annotate the video frames automatically. Our method leverages the semantic feature extraction of our framework presented in Section 3.1.

In the first part of our algorithm, we assume we receive a set of unlabeled video sequences (without any annotation in both training and testing phases), we first extract the labels of each video frame using our Semantic feature extraction method presented in Section 3.1. We apply semantic feature extraction of our framework on each observed frame. The output of BRNN layer gives the set of objects and actions that are seen in each video frame (Stage 1 Figure 2). Then we use a pre-defined lookup table, which has a list of actions that are used in training BRNN in the Semantic feature extraction section 3.1.2, and the output of BRNN layer in order to label our unlabeled video sequences (Stage 2 Figure 2). Once we know the frame labels, we train our classifier using the same method as presented in Section 3 and Section 4 (Stage 3 Figure 2). We note that in the labeling phase we do not have any information about the number of actions or the name of actions that we expect to label in the observed videos and we leave our method to annotate the videos automatically.

## 5. Experiments

This section describes our experiments on three publicly available databases, MPII Cooking ActivitiesRohrbach et al. (2012), UCF 101-Action RecognitionSoomro et al. (2012) and Hollywood Laptev et al. (2008). We chose these datasets because they are challenging and they consist complex scenes. In addition, the included video clips have a small number of positive training examples which can challenge the robustness of our algorithm against a small number of training examples. We note that, since one of the contributions of this paper is to tackle the problem of complex event detection on the continues domain, we utilize the videos from UCF 101-Action Recognition and Hollywood datasets (which both are challenging datasets) in order to create long lasting videos with multiple actions with unknown starting and ending locations. We define our protocol in the subsequent subsections.

An overview of the databases is presented in Section ??; Section 4.3 details the experimental settings used and Section 4.4 presents our results for event detection tasks; and Section 4.5 compares our proposed approach with other state-of-the-art methods.

### 5.1. Datasets

**MPII Cooking Activities dataset:** The MPII cooking dataset contains 65 different cooking activities performed by 12 participants. This database is composed of videos recorded with a 4D View Solutions system using a Point Grey Grashopper camera with $1226 \times 1224$ pixel resolution at 24.4 fps and global shutter. The camera is attached to the ceiling, recording a person working at the counter from the front and preparing a dish. In total this dataset contains 44 videos with a total length of more than 8 hours and 881,755 frames. This dataset includes long videos and is a challenging dataset due to the high number of classes. In order to test our algorithm on this dataset we choose 16 different classes for evaluation. We discard the events which are either too elementary and simple to form a composite activity (e.g. how to secure a chopping board), or were duplicated with slightly different titles. For evaluation we chose 80% of the videos for the training and the 20% for testing.Some examples from this dataset are shown in Figure 4.

**UCF 101-Action Recognition dataset:** This dataset is a challenging action recognition dataset and contains 13, 320 videos from 101 action categories collected directly from YouTube. The videos are in $320 \times 240$ pixels resolution and 25 frames per second. Since each video has only one action, in order to evaluate our continuous event detection method on this dataset we consider detecting the actions: "BaseballPitch", "Basketball", "BasketballDunk", "Bowling", "CliffDiving", "CricketShot", "CuttingInKitchen" in a long duration video by concatenating multiple videos. In particular, each video contains one action of interest which is proceeded and succeeded by six different actions. The position of the action of interest is randomly chosen and the starting and ending locations of the action are unknown. The generated dataset consists in total 105 videos. For evaluation we generate 70 training and 35 testing videos from this dataset. Figure 4 illustrates some examples from this dataset.

**Hollywood dataset:** Hollywood dataset contains videos with human action from 32 movies and is composed of 8 action classes: "AnswerPhone", "GetOutCar", "HandShake", "HugPerson", "Kiss", "SitDown", "SitUp", "StandUp". In order to generate long duration and challenging videos with multiple actions we concatenate one action of interest with six different actions similar to the protocol we use for UCF 101 dataset. The generated dataset consists in total 32 videos. The dataset is divided into a test set obtained from 20 movies and training set obtained from 12 movies different from the test set. Figure 4 shows some examples from this dataset.

### 5.2. Experimental setup

**Evaluation Metrics:** In order to report the performance of our proposed model we report the Average Precision metric (AP) because it is a better measure for event detection:

$$precision = \frac{|\mathbf{y} \bigcap \mathbf{y}^*|}{|\mathbf{y}|}$$

where $\mathbf{y}$ is the detector output at time $t$ (end of window) and $\mathbf{y}^*$ is the ground truth.

**Number of temporal codebooks:** For building the codebooks, k-means clustering is used. In our experiments we perform cross-validation on training data to tune the number of temporal codebooks in Eq. 4.

### 5.3. Results

This section provides the results on the proposed datasets. As we mentioned earlier state-of-the-art methods utilize local features from the local spatial or spatio-temporal patches from the video frames for event detection. In this work in order to gain further insight into the performance of our proposed model against other algorithms we consider three different feature representation techniques: (i)"*Improved Dense Trajectories (IDT)*", (ii) "*Optical Flow*" and (iii) "*Convolutional Neural Network (CNN)*" as the video encoding methods. To extract IDT features we use the model explained in Section 3.2 and for optical flow features we use the algorithm by

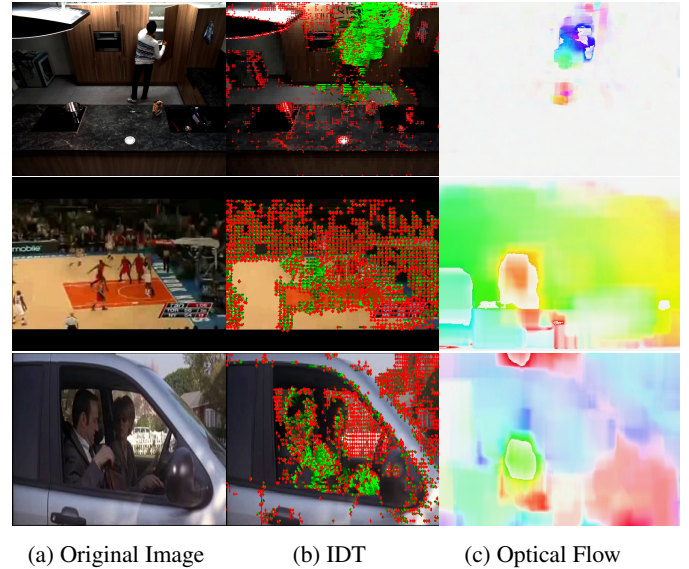

(a) Original Image    (b) IDT    (c) Optical Flow

Fig. 3: Examples of different feature extraction methods on different datasets. (a) the original image from the dataset. (b) IDT features between two consecutive frames. (c) Optical flow features between two consecutive frames.

Färnback Farnebäck (2003) as implemented in the OpenCV library [1] and we represent Optical flow features as Histograms of Optical Flow (HOF). In order to construct the video descriptors we use the standard bag-of-feature method introduced in Eq. 4 and for generating codebooks we use k-means clustering. Some examples from the IDT and optical flow features extracted from the video frames can be found in Figure 3. Finally for the CNN features we use the model presented in Section 3.1.1 (the $4096 \times 1$ vector extracted from the last fully connected layer from the 16 layers CNN architecture).

As explained in Section 3 our technique utilizes the combination of video descriptors using IDT and semantic features for event detection. In order to extract the semantic features we use the algorithm explained in Section 3.1 and apply it to each video frame. Some examples of the extracted features and their corresponding input images are shown in Figure 4.

Table 1, Table 2 and Table 3 report the performance of our proposed model introduced in Section 4.1 against different feature encoding methods. For evaluation, in order to compare our model against other feature representation techniques we use the same formulation as presented in Eq. 7. In this presentation we denote the features captured by $\psi_S(.)$ as the primary features, and those captured by $\psi_T(.)$ as the secondary features. In these tables "CNN + Semantic" corresponds to the combination of CNN (as the primary) and Semantic (as the secondary) features for Eq. 7. "IDT + CNN" corresponds to the mixture of dense trajectories (as the primary) and CNN (as the secondary) features. For "CNN", "IDT" and "Optical Flow" methods we use exactly the same features for both the primary and secondary features. "IDT + Semantic" corresponds to our proposed model in which IDT features are utilized as the primary and Semantic features are used as the secondary features. From

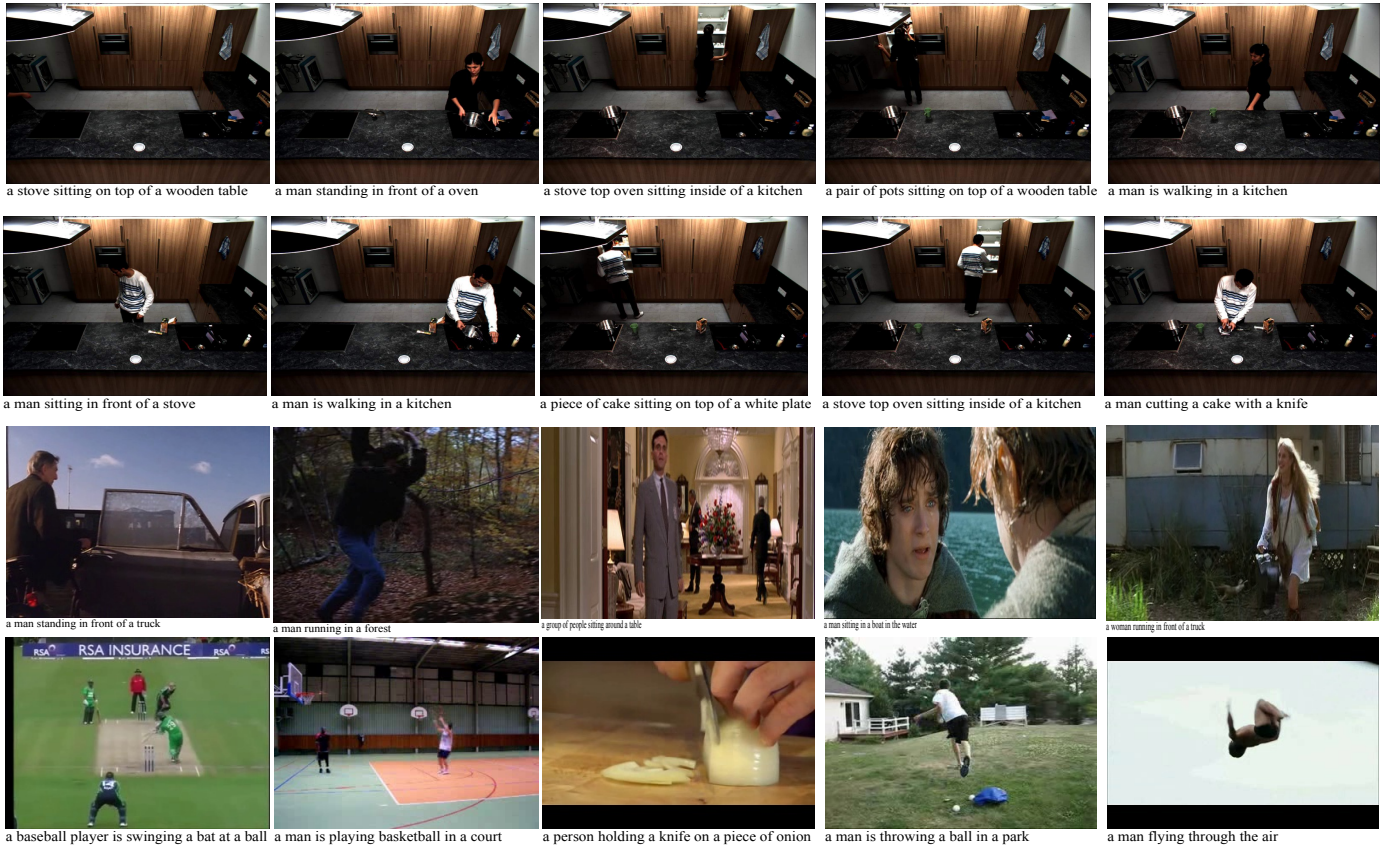

Fig. 4: Some examples of the semantic features extracted from each video frame using the method presented in Section 3.1.

the results we can see the impact of different feature representation techniques on each of the event categories. As can be seen from the tables, on all datasets IDT features perform better than CNN. On the other hand, CNN outperforms optical flow. In addition, we can see how mixing the other features to the model can improve the detection performance. From the tables we can see our proposed model "IDT + Semantic" performs significantly better than all the other techniques. It is worth mentioning that for the events such as "Takeout/Drawer" and "Takeout/Cupboard", "Open/Drawer" and "Open/Cupboard" the semantic features improve the detection performance, however other techniques fail to detect these events accurately.

We also use our model for continuous event detection on unlabeled videos using the method presented in Section 4.2. We use the same video sequences as explained in Section **??**. We note that in this experiment we do not have any prior knowledge about the frame labels in the both training and testing phases. As explained in Section 4.2 we first feed the training sequences to our semantic feature extraction method (Section 3.1) in order to automatically annotate the training video frames. Once we are finished with the labeling phase, then we train our classifier (Section 4) using the annotated frames. After training our classifier we test our model on the unseen videos. Figure 5 shows the results obtained for this experiment in comparison with the ground truth labels. As can be seen from the figure, for the events such as "Bckgrd Activity" and "Basketball Dunk" the auto-labeling is not performed efficiently. One reason is that, because for "Bckgrd Activity" in each video frame the semantic feature extraction technique observes a set of objects and actions, therefore the model fails to accurately label these frames as the "Bckgrd Activity".

Figure 6 compares the supervised against unsupervised classification results. By supervised we mean that we use the ground truth labels as the training labels while we are training our classifier and by unsupervised we mean the method which is presented in Section 4.2. From the results we can see that the unsupervised classification performs slightly less than supervised classification method. However, the supervised classification technique is limited by the availability of the training labels. On the other hand, our method is able to generalize to any arbitrary dataset.

### 5.4. Comparison with other methods

We also compare our approach against other techniques. Since there is a little work in the literature on continues event detection, we consider two approaches for comparison, Abbasnejad et al. Abbasnejad et al. (2015) and Hoai et al.Hoai and De la Torre (2014). In Abbasnejad et al. (2015) they use the idea of sliding windows in a max margin framework for continuous event detection and in Hoai and De la Torre (2014) they utilize the SO-SVM in conjunction with BoW for early event detection. The results for this comparison is shown in Tabel 4. As can be seen from the table our method outperforms other approaches.

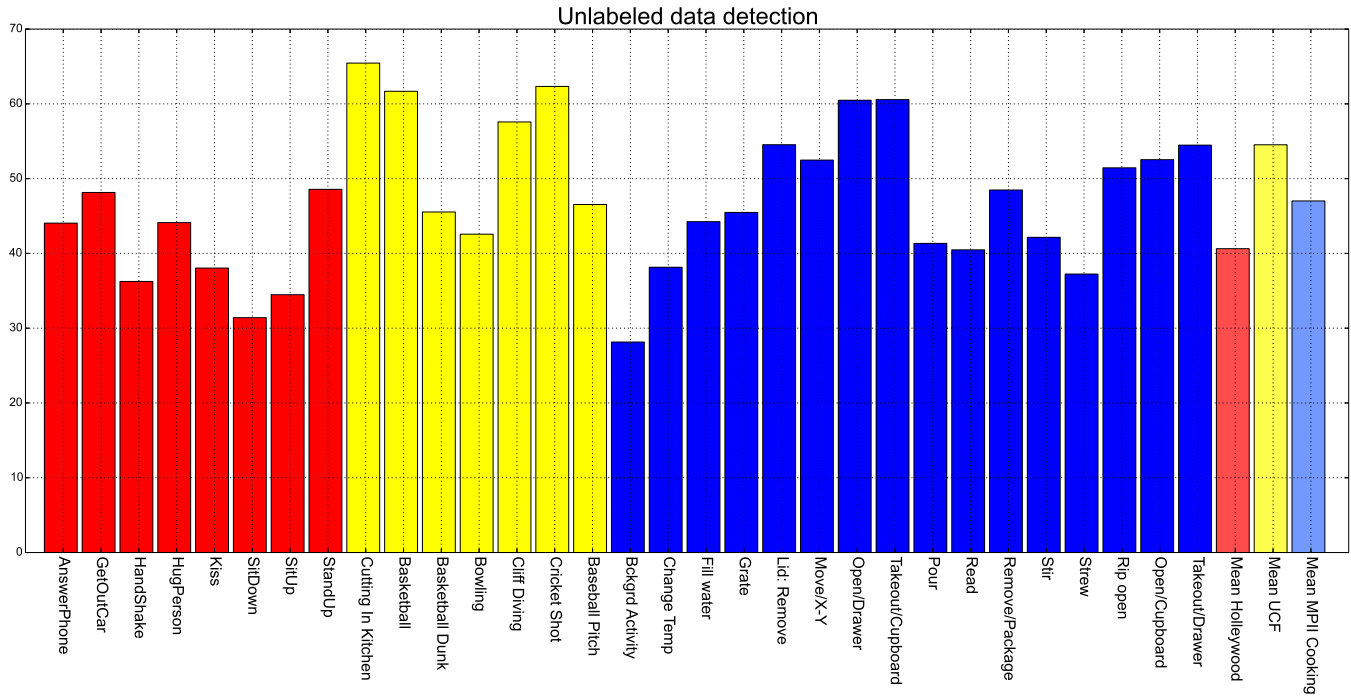

Fig. 5: Results for the experiment on the unlabeled data.

Table 1: Results on the MPII cooking activities dataset.

| Event Class | CNN + Semantic | Semantic | Optical Flow | IDT | CNN | IDT + CNN | IDT + Semantic |
|---|---|---|---|---|---|---|---|
| Bckgrd Activity | 49.32 | 28.15 | 25.74 | 36.58 | **57.58** | 56.27 | 52.42 |
| Change Temp | 52.20 | 38.15 | 31.85 | 37.46 | 41.64 | 52.02 | **56.21** |
| Fill water | 56.53 | 44.25 | 53.41 | 60.91 | 61.17 | 62.11 | **63.06** |
| Grate | 40.70 | 45.48 | 31.81 | 35.23 | 34.29 | 41.29 | **51.61** |
| Lid: Remove | 35.94 | 54.54 | 33.28 | **38.88** | 36.13 | 34.16 | 35.68 |
| Move X/Y | 37.09 | 52.48 | 32.57 | 51.39 | 40.76 | 50.64 | **51.88** |
| Open/Drawer | 51.24 | 60.48 | 32.09 | 54.39 | 50.21 | 62.23 | **64.20** |
| Take-out/Cupboard | 68.29 | 60.57 | 58.90 | 69.93 | 68.98 | 70.21 | **71.91** |
| Pour | 32.01 | 41.34 | 30.75 | 44.12 | 31.70 | 42.87 | **46.72** |
| Read | 34.27 | 40.48 | 31.83 | 39.81 | **42.39** | 39.80 | 40.55 |
| Re-move/Package | 39.12 | 48.48 | 28.41 | 38.92 | 38.27 | 33.63 | **67.21** |
| Stir | 49.83 | 42.15 | 38.16 | **56.79** | 50.21 | 47.92 | 48.19 |
| Strew | 41.33 | 37.24 | 35.87 | 40.86 | 40.74 | 40.42 | **44.73** |
| Rip Open | 32.17 | 51.45 | 30.09 | 44.65 | 29.37 | 41.52 | 39.21 |
| Open/Cupboar | 46.42 | 52.54 | 33.36 | 39.14 | 44.21 | 50.21 | **62.13** |
| Take-out/Drawer | 51.17 | 54.48 | 41.19 | 51.52 | 52.09 | 54.76 | **64.42** |
| **Mean** | 49.91 | 47.01 | 34.83 | 45.97 | 44.98 | 50.00 | **55.07** |

Table 2: Results on the UCF 101-Action Recognition dataset.

| Event Class | CNN + Semantic | Semantic | Optical Flow | IDT | CNN | IDT + CNN | IDT + Semantic |
|---|---|---|---|---|---|---|---|
| Cutting In Kitchen | 67.32 | 65.45 | 44.74 | 46.58 | 65.58 | 77.27 | **78.42** |
| Basketball | **68.20** | 61.68 | 41.35 | 47.46 | 42.64 | 64.42 | 66.28 |
| Basketball Dunk | 65.71 | 45.54 | 55.81 | **69.91** | 68.61 | 67.16 | 69.72 |
| Bowling | **66.98** | 42.57 | 48.81 | 56.23 | 61.29 | 64.52 | 66.92 |
| Clift Diving | 54.94 | 57.58 | 38.28 | 45.88 | 38.13 | 40.68 | **59.68** |
| Cricket Shot | 61.09 | 62.33 | 40.57 | 51.39 | 47.76 | **69.89** | 69.76 |
| Baseball Pitch | 66.91 | 46.54 | 56.09 | **70.39** | 57.21 | 66.43 | 69.20 |
| **Mean** | 64.45 | 54.52 | 46.52 | 55.41 | 54.61 | 64.20 | **68.57** |

### 5.5. Discussion and limitations

As mentioned earlier the main aim of this paper is to tackle the problem of complex event detection in the continuous domain. Our method combines both semantic and temporal features in an adaptive framework, and we demonstrate that combination of different sets of features (i.e. set of objects, actions and temporal information) improves the detection accuracy. Semantic features are used to extract the information about the objects and actions that are seen in each video frame, while temporal features are utilized to capture the temporal pattern properties in the observed sequences. This combination increases the detection performance dramatically, specially for the events such as "Takeout/Drawer" and "Takeout/Cupboard", "Open/Drawer" and "Open/Cupboard". On the other hand, we can see that "Optical Flow" features perform poorly on all the three datasets, since they only carry the temporal properties and complex event detection based on temporal information is insufficient. Using "IDT" features performs better than "CNN" and "Semantic" features. However, combining "IDT" with "CNN" features ("IDT+CNN") improves the detection performance by almost 6%. Combining "IDT" and "Semantic" features ("IDT+Semantic") performs the best among all the datasets, since it combines the temporal properties as well as the information about the objects. We also show that our method can be used to automatically annotate unlabeled videos, where we do not have any prior knowledge about their labels.

| Event Class | CNN + Semantic | Semantic | Optical Flow | IDT | CNN | IDT + CNN | IDT + Semantic |
|---|---|---|---|---|---|---|---|
| Answer-Phone | 48.02 | 44.05 | 34.43 | 38.23 | 39.19 | 48.43 | **51.76** |
| GetOutCar | 49.18 | 48.15 | 35.76 | 41.98 | 39.65 | **51.16** | 50.62 |
| HandShake | 40.19 | 36.25 | 35.65 | 38.32 | 37.76 | **42.16** | 41.31 |
| HugPerson | 43.54 | 44.12 | 37.07 | 43.17 | 40.22 | **44.56** | 44.26 |
| Kiss | 54.72 | 38.04 | 49.32 | 52.38 | 51.42 | **56.86** | 54.41 |
| SitDown | 44.24 | 31.41 | 42.29 | 45.21 | 48.61 | 47.25 | **51.76** |
| SitUp | 33.09 | 34.48 | 31.14 | 34.87 | 31.47 | 35.46 | **37.49** |
| StandUP | 52.27 | 48.57 | 46.08 | 55.87 | 50.11 | 56.43 | **57.05** |
| Mean | 45.66 | 40.63 | 38.98 | 43.75 | 42.30 | 47.79 | **48.58** |

Table 3: Results on the Hollywood dataset.

| Method | MPII cooking | UCF Action | Hollywood |
|---|---|---|---|
| Proposed approach | **55.07** | **68.57** | **48.58** |
| Abbasnejad et al. Abbasnejad et al. (2015) | 54.27 | 65.08 | 47.46 |
| Hoai et al. Hoai and De la Torre (2014) | 51.13 | 64.41 | 46.84 |

Table 4: Comparison of our results to the state of art.

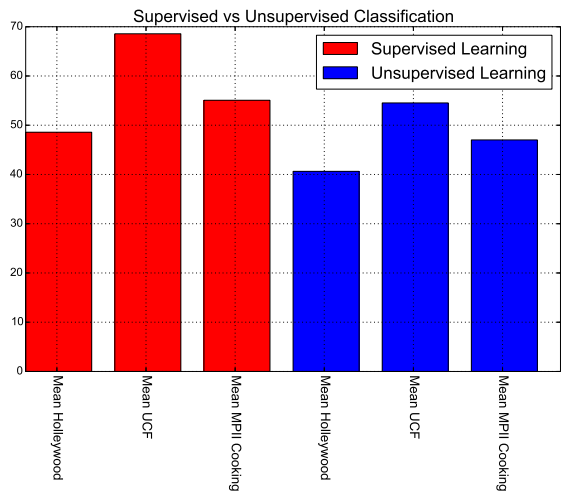

Fig. 6: Comparison between the supervised and unsupervised classification methods.

Although our method performs well, but there are a few limitations with it as well. One limitation is that, despite that our method can be applied to different action and event datasets, unlike the methods presented in Abbasnejad et al. (2015); Hoai and De la Torre (2014) we cannot apply our algorithm to different temporal problems such as facial expression analysis or gesture recognition. Moreover, when we have strong temporal information and unrelated objects in the background (some activities such as playing basketball) our method cannot perform efficiently.

## 6. Conclusion and Future work

In this paper the problem of event detection in the complex scenes is addressed. Previous methods have demonstrated that combination of multiple sets of features can improve the detection performance dramatically. A drawback to these methods is that they are only limited to the videos with specific properties (i.e. videos with audio and text subtitles), and it is hard to generalize them to any arbitrary input video. In this work we introduce a novel approach based on the combination of temporal and semantic features. Our framework enables us to model the sequential frame-by-frame videos using the sets of the objects and actions that are extracted in each arrival frames. This approach proved effective in our empirical evaluations on three challenging datasets. We also show that our proposed approach can be extended to the unlabeled videos, where we do not have any prior information about the videos' labels. We utilize our semantic feature extraction of our method and label each arrival frame based on the extracted actions.

On the other hand, one direction for the future work would be a more complete analysis of the set of objects that are extracted as semantic features in each frame and their effect on the classification accuracy. Furthermore, using an iterative algorithm to improve the performance of the unlabeled event detection accuracy. By training a model on the unlabeled videos and using the classification outputs of the train model and retraining the new model based on the new results and repeating until convergence. We can also try to predict the false classification events by drawing the correlation between the surrounding objects and the proceeding ad succeeding events.